\documentclass[runningheads]{llncs}
\usepackage{graphicx}
\usepackage{multirow}
\usepackage{booktabs}
\usepackage{bbding}
\usepackage{hyperref, color}
\usepackage{amsmath}
\usepackage{algorithm}
\usepackage{algorithmic}
\usepackage[switch]{lineno}
\usepackage{threeparttable}
\usepackage{rotating}

\begin{document}
\title{More than Segmentation: Benchmarking \\ SAM 3 for Segmentation, 3D Perception, \\ and Reconstruction in Robotic Surgery}

\titlerunning{More than Segmentation: Benchmarking SAM 3 for Segmentation, 3D Perception and Reconstruction in Robotic Surgery}
\author{Wenzhen Dong\inst{1}
\and Jieming Yu\inst{2}
\and Yiming Huang\inst{1}
\and Hongqiu Wang\inst{3}
\and Lei Zhu\inst{3}
\and Albert C. S. Chung\inst{2}
\and Hongliang Ren\inst{1} 
\and Long Bai\inst{4}
}

\authorrunning{W. Dong et al.}


\institute{The Chinese University of Hong Kong
\and The Hong Kong University of Science and Technology
\and The Hong Kong University of Science and Technology (Guangzhou)
\and Technical University of Munich
}

\maketitle             
\begin{abstract}
The recent SAM 3 and SAM 3D have introduced significant advancements over the predecessor, SAM 2, particularly with the integration of language-based segmentation and enhanced 3D perception capabilities. SAM 3 supports zero-shot segmentation across a wide range of prompts, including point, bounding box, and language-based prompts, allowing for more flexible and intuitive interactions with the model. In this empirical evaluation, we assess the performance of SAM 3 in robot-assisted surgery, benchmarking its zero-shot segmentation with point and bounding box prompts and exploring its effectiveness in dynamic video tracking, alongside its newly introduced language prompt segmentation. While language prompts show potential, their performance in the surgical domain is currently suboptimal, highlighting the need for further domain-specific training. Additionally, we investigate SAM 3D's depth reconstruction abilities, demonstrating its capacity to process surgical scene data and reconstruct 3D anatomical structures from 2D images. Through comprehensive testing on the MICCAI EndoVis 2017 and EndoVis 2018 benchmarks, SAM 3 shows clear improvements over SAM and SAM 2 in both image and video segmentation under spatial prompts, while the zero-shot evaluations of SAM 3D on SCARED, StereoMIS, and EndoNeRF indicate strong monocular depth estimation and realistic 3D instrument reconstruction, yet also reveal remaining limitations in complex, highly dynamic surgical scenes.
% \keywords{}
\end{abstract}
\section{Introduction}
\label{sec:introduction}

%%%%%%%%%%%%%%%%%% Intro %%%%%%%%%%%%%%%%%%%%%%%%%%%%
Artificial Intelligence (AI) in surgery is increasingly vital for improving various aspects of surgical workflows~\cite{zia2023surgical,bai2023surgical,hao2025enhancing,wang2023dynamic}. Surgical segmentation is fundamental in robotic surgery, as it directly impacts real-time guidance, motion tracking, and the precision of automated interventions~\cite{liu2025resurgsam2,liu2025sam2s,wang2024video,yu2024adapting,yuan2024segment}. The complex and dynamic nature of surgical scenes, characterized by intricate anatomical structures, occlusions, and fast-moving instruments, requires robust and adaptive segmentation techniques. In addition to segmentation, 3D perception plays a critical role in providing a comprehensive understanding of the surgical environment, allowing for better spatial awareness and instrument tracking in three dimensions~\cite{huang2025surgtpgs,huang2024endo,zhu2024endogs}. This is particularly important in surgeries where depth, orientation, and precise instrument positioning are crucial for success~\cite{huang2025advancing,ozyoruk2021endoslam}.

Building on the success of SAM 2~\cite{ravi2024sam}, SAM 3~\cite{carion2025sam} and SAM 3D~\cite{chen2025sam} introduce two key advancements: language-based segmentation and 3D perception. SAM 3 enables segmentation through natural language prompts, allowing users to specify segmentation targets using verbal descriptions. While the performance of language-based segmentation in surgical applications is still under exploration, it holds significant potential for reducing the reliance on predefined categories or visual prompts. SAM 3D also brings the ability to generate 3D reconstructions from 2D images, offering a more comprehensive understanding of the surgical environment. This feature is particularly valuable in complex surgical tasks, where the ability to perceive and track instruments in three dimensions enhances both precision and decision-making.

In this empirical evaluation, we explore the newly introduced capabilities of SAM 3 and SAM 3D, including language-based segmentation and 3D perception. We examine how these advancements can enhance robotic surgery, particularly in dynamic and complex surgical environments. While these new features are the focus of our exploration, we also assess the performance of traditional point and bounding box prompt segmentation methods to provide a comprehensive comparison and understand the broader impact of the SAM series in surgical applications.
Our findings indicate that while SAM 3 and SAM 3D shows promise, particularly with its 3D perception abilities, there are areas for improvement, especially with language-based prompts in the surgical context. Further adaptation and training will be necessary to fully realize its potential in surgical applications. Specifically, our contributions and findings can be summarized as:

\begin{itemize}
    \item We benchmark SAM 3 on standard robotic surgery datasets in both image and video settings with point, box, and text prompts. In the zero-shot setting, SAM 3 with box prompts achieves new state-of-the-art segmentation performance on EndoVis17 and EndoVis18 and generally outperforms SAM and SAM 2 variants in both binary and instrument-level segmentation.
    \item We systematically analyze different prompting modalities for SAM 3 in surgical scenes, including its new language-based (text) prompts. While spatial prompts (point/box) remain strong, text prompts perform poorly in capturing surgical semantics, revealing a clear domain gap but also a promising direction for surgery-specific adaptation.
    \item We present the zero-shot evaluation of SAM 3D for monocular depth estimation in endoscopic surgery, where SAM 3D achieves lower depth errors and higher accuracy than several prior methods trained or fine-tuned on endoscopic data, highlighting the strength of its pretrained 3D priors despite slower inference.
    \item We further investigate SAM 3D for 3D surgical instrument segmentation and reconstruction on StereoMIS and EndoNeRF, using a 3D evaluation protocol based on reprojection to labeled point clouds. SAM 3D recovers plausible 3D instrument geometry and performs reasonably in static-camera settings, while failures in highly dynamic, narrow-baseline sequences expose important limitations for future 3D surgical perception.
\end{itemize}

\begin{table}[h]
  \centering
  \caption{Quantitative comparison of binary and instrument segmentation on EndoVis17 and EndoVis18 datasets. For SAM 2 and SAM 3, we present results in images and videos.}
  \begin{threeparttable}
    \resizebox{\textwidth}{!}{
    \begin{tabular}{c|lcc|cc|cc}
    \toprule
    \multirow{2}{*}{Type} & \multirow{2}{*}{Method} & \multirow{2}{*}{Pub/Year (20-)} & \multirow{2}{*}{Arch.} & \multicolumn{2}{c}{EndoVis17} & \multicolumn{2}{c}{EndoVis18} \\
    \cline{5-8}
    & & & & Binary IoU & Instrument IoU & Binary IoU & Instrument IoU\\
    \hline
    \multirow{6}{*}{Single-Task}
      & Vanilla UNet~\cite{ronneberger2015u} & MICCAI15 & UNet & 75.44 & 15.80 & 68.89 & - \\
      & TernausNet~\cite{shvets2018automatic} & ICMLA18 & UNet & 83.60 & 35.27 & - & 46.22 \\
      & MF-TAPNet~\cite{jin2019incorporating} & MICCAI19 & UNet & 87.56 & 37.35 & - & 67.87 \\
      & Islam et al.~\cite{islam2019real} & RA-L19 & - & 84.50 & - & - & - \\
      & ISINet~\cite{gonzalez2020isinet} & MICCAI21 & Res50 & - & 55.62 & - & 73.03 \\
      & Wang et al.~\cite{wang2022rethinking} & MICCAI22 & UNet & - & - & 58.12 & - \\
    \hline
    \multirow{5}{*}{Multi-Task}
      & ST-MTL~\cite{islam2021st} & MedIA21 & - & 83.49 & - & - & - \\
      & AP-MTL~\cite{islam2020ap} & ICRA20 & - & 88.75 & - & - & - \\
      & S-MTL~\cite{seenivasan2022global} & RA-L22 & - & - & - & - & 43.54 \\
      & TraSeTR~\cite{zhao2022trasetr} & ICRA22 & Res50+Trfm & - & 60.40 & - & 76.20 \\
      & S3Net~\cite{baby2023forks} & WACV23 & Res50 & - & 72.54 & - & 75.81 \\
    \hline
    \multirow{11}{*}{Prompt-based *}
      & SAM (1 Point)~\cite{kirillov2023segment} & ICCV23 & ViT\_h & 53.88 & 55.96 & 57.12 & 54.30 \\
      & SAM (Box)~\cite{kirillov2023segment} & ICCV23 & ViT\_h & 89.19 & 88.20 & 89.35 & 81.09 \\
      & SAM 2-Image (1 Point)~\cite{ravi2024sam} & ICLR25 & ViT\_h & 84.96 & 81.10 & 77.14 & 73.76 \\
      & SAM 2-Image (Box)~\cite{ravi2024sam} & ICLR25& ViT\_h & 90.97 & 86.92 & 90.18 & 81.97 \\
      & SAM 2-Video (1 Point)~\cite{ravi2024sam} & ICLR25 & ViT\_h & 62.45 & 58.74 & 65.19 & 57.59 \\
      \cline{2-8}
      & SAM 3-Image (1 Point)~\cite{carion2025sam} & arXiv25 & ViT\_h & 79.15 & 77.57 & 82.43 & 81.14 \\
      & SAM 3-Image (Box)~\cite{carion2025sam} & arXiv25 & ViT\_h & \textbf{92.61} & \textbf{91.67} & \textbf{93.33} & \textbf{91.88} \\
      & SAM 3-Image (Text)~\cite{carion2025sam} & arXiv25 & ViT\_h & 20.07 & 9.09 & 48.49 & 15.99 \\
      & SAM 3-Video (1 Point)~\cite{carion2025sam} & arXiv25 & ViT\_h & 82.07 & 73.94 & 54.87 & 44.00 \\
      & SAM 3-Video (Box)~\cite{carion2025sam} & arXiv25 & ViT\_h & 61.48 & 61.10 & 77.43 & 71.58 \\
      & SAM 3-Video (Text)~\cite{carion2025sam} & arXiv25 & ViT\_h & 6.16 & 2.11 & 14.43 & 17.00 \\
    \bottomrule
    \end{tabular}}
  \end{threeparttable}
  \begin{tablenotes}
    \footnotesize
    \item{*} Categorical information directly inherits from associated prompts.
  \end{tablenotes}
  \label{tab:seg_res}
\end{table}

\section{2D Instruments Segmentation with Prompts}
\subsection{Implementation Details}

\noindent \textbf{Datasets.}
We conduct the empirical evaluation on the EndoVis17~\cite{allan20192017} and EndoVis18~\cite{allan20202018} datasets, following the validation splits introduced in~\cite{wang2023sam}. Both datasets provide video data for surgical instrument segmentation, with segmentation annotations at a rate of 1 frame per second.
\\
\\
\noindent \textbf{Comparison Methods.}
Baseline methods follow the configuration described in~\cite{yu2024sam}, and their performance is summarized in Table~\ref{tab:seg_res}. 
Please note that a fully fair comparison is difficult. Traditional instrument segmentation methods are based on supervised learning and do not require prompts during inference. In contrast, the SAM series models rely on user-provided cues and are evaluated in a zero-shot setting.
\\
\\
\noindent \textbf{Prompting Setup.}
We evaluate surgical instrument segmentation using three prompt types (point, bounding box, and text) under static-image and video settings. For image inputs, prompts are generated directly from ground-truth annotations. For video sequences, prompts are provided \emph{only on the first frame} of each sequence, and the model is required to propagate segmentation to all subsequent frames without further user intervention. All SAM models operate in a zero-shot manner. Specifically, for each instrument instance, we construct the prompts as follows: 
\textbf{(i) 1-Point Prompt:} a single foreground point located at the centroid of its ground-truth mask; 
\textbf{(ii) Box Prompt:} a tight axis-aligned bounding box enclosing the instrument mask;
\textbf{(iii) Text Prompt:} the instrument category name (e.g., \emph{Bipolar Forceps}, \emph{Prograsp Forceps}), used as semantic guidance without spatial supervision.

\begin{figure*}[t]
  \centering
  \includegraphics[width=\linewidth, trim=10 20 10 0]{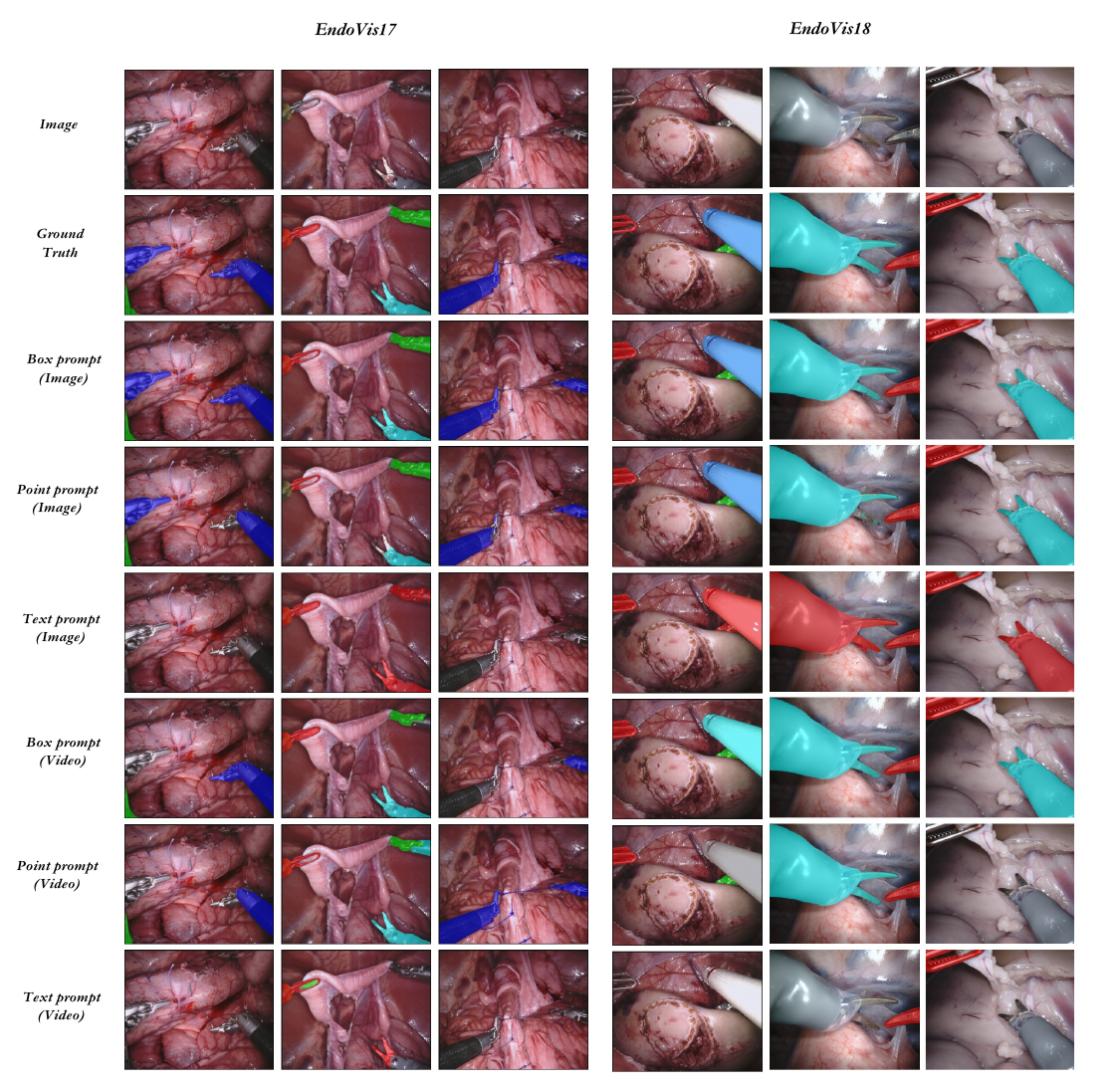}
  \caption{Qualitative results of SAM 3 on 6 images from EndoVis17 and EndoVis18 on 6 different tasks.}
  \label{fig:2d_seg}
\end{figure*}

\subsection{Results and Analysis}

The quantitative and qualitative results of the two datasets are shown in Fig.~\ref{fig:2d_seg} and Table~\ref{tab:seg_res}. Overall, SAM 3 shows clear improvement over its predecessors in image and video segmentation. However, in two specific setups, SAM 3 performs slightly worse than SAM 2: image segmentation on EndoVis17 using 1-point prompts, and video segmentation on EndoVis18 using 1-point prompts. This may be due to the domain gap between surgical and natural images, which reduces model generalization.
Additionally, different prompting methods also have different effects on the segmentation results. For image segmentation, box prompts consistently provide better results than 1-point prompts. For video segmentation, using box prompts with SAM 3 did not lead to stable performance improvement. The effectiveness of box prompts may depend on dataset-specific traits such as scene layout, instrument appearance, or surgical motion. While box prompts can help in some cases, their potential added complexity may not always suit a given dataset or task, thus producing mixed outcomes.

SAM 3 also adds text prompts for image and video segmentation. However, it shows clear limits in understanding surgical scene semantics and performs poorly with text prompts. Although text prompts could be more useful in clinical practice, the current SAM 3 lacks the surgical knowledge and the broad image-text training needed to make this feature work well. Further fine-tuning on clinical datasets is needed to make the model clinically useful for surgery.

\section{Depth Map Reconstruction}

\subsection{Implementation Details}
\noindent \textbf{Datasets \& Evaluation Metrics.}
We evaluate our method on the SCARED dataset~\cite{allan2021stereo}, following the validation set partitioning in~\cite{recasens2021endo}. The SCARED dataset consists of static endoscopic scenes. Following the approach in the literature~\cite{guo2025endo3r}, we use six metrics to evaluate the depth estimation performance and efficiency: Abs Rel, Sq Rel, RMSE, log RMSE, $\delta < 1.25$, and FPS.
\\
\\
\noindent \textbf{Comparison Methods.}
The baseline methods follow the configuration in~\cite{guo2025endo3r}, including depth estimation methods for general scenarios and those specifically designed for surgical scenarios. All methods were fine-tuned for the endoscopic domain, as described in~\cite{guo2025endo3r}. 
The experiments are conducted on the NVIDIA RTX A6000 GPU.

\begin{table}[t]
\scriptsize
\centering
\caption{Depth estimation performance on the SCARED dataset.}
\vspace{-1em}
\label{tab:depth_results}
\resizebox{\linewidth}{!}{%
\begin{tabular}{@{}l|ccccc|c@{}}
\toprule
\textbf{Methods}&\textbf{Abs Rel}$\downarrow$ & \textbf{Sq Rel}$\downarrow$ & \textbf{RMSE}$\downarrow$ & \textbf{RMSE log}$\downarrow$&\textbf{$\boldsymbol{\delta}<\text{1.25}\uparrow$}  & \textbf{FPS}$\uparrow$
\\ 
\midrule
Monodepth2~\cite{godard2019digging} & 0.432& 3.548&4.704&0.431&0.425&22.05\\
Endo-SfM~\cite{ozyoruk2021endoslam} & 0.241& 0.865&2.286&0.267&0.585&7.33 \\
AF-SfM~\cite{shao2022self} & 0.257& 0.960&2.162&0.291&0.573&3.17\\ 
EndoDAC~\cite{cui2024endodac} & 0.242& 0.934&2.014&0.275&0.584&31.79\\
Transfer~\cite{budd2024transferring} &0.297& 1.207&2.561&0.319&0.561&9.37\\
DA-V2~\cite{yang2025depth} & 0.313& 1.425&2.839&0.453&0.508&4.18\\
VDA~\cite{chen2025video} & 0.291& 1.186&2.447&0.296&0.647&6.86\\
Endo DM~\cite{recasens2021endo}& 0.203& 0.651&2.063&0.245&0.612&14.58 \\
Monst3R~\cite{zhang2024monst3r}& 0.198& 0.539&1.965&0.234&0.626&18.68 \\
Endo3R~\cite{guo2025endo3r} & 0.124 & 0.227 & 1.209 & 0.135 &0.839 &19.17\\
SAM 3D~\cite{chen2025sam} & 0.072 & 0.089 & 0.849 & 0.093 &  0.957 &0.16\\
\bottomrule 
\end{tabular}}
\end{table}

\begin{figure*}[t]
  \centering
  \includegraphics[width=\linewidth, trim=0 60 0 0]{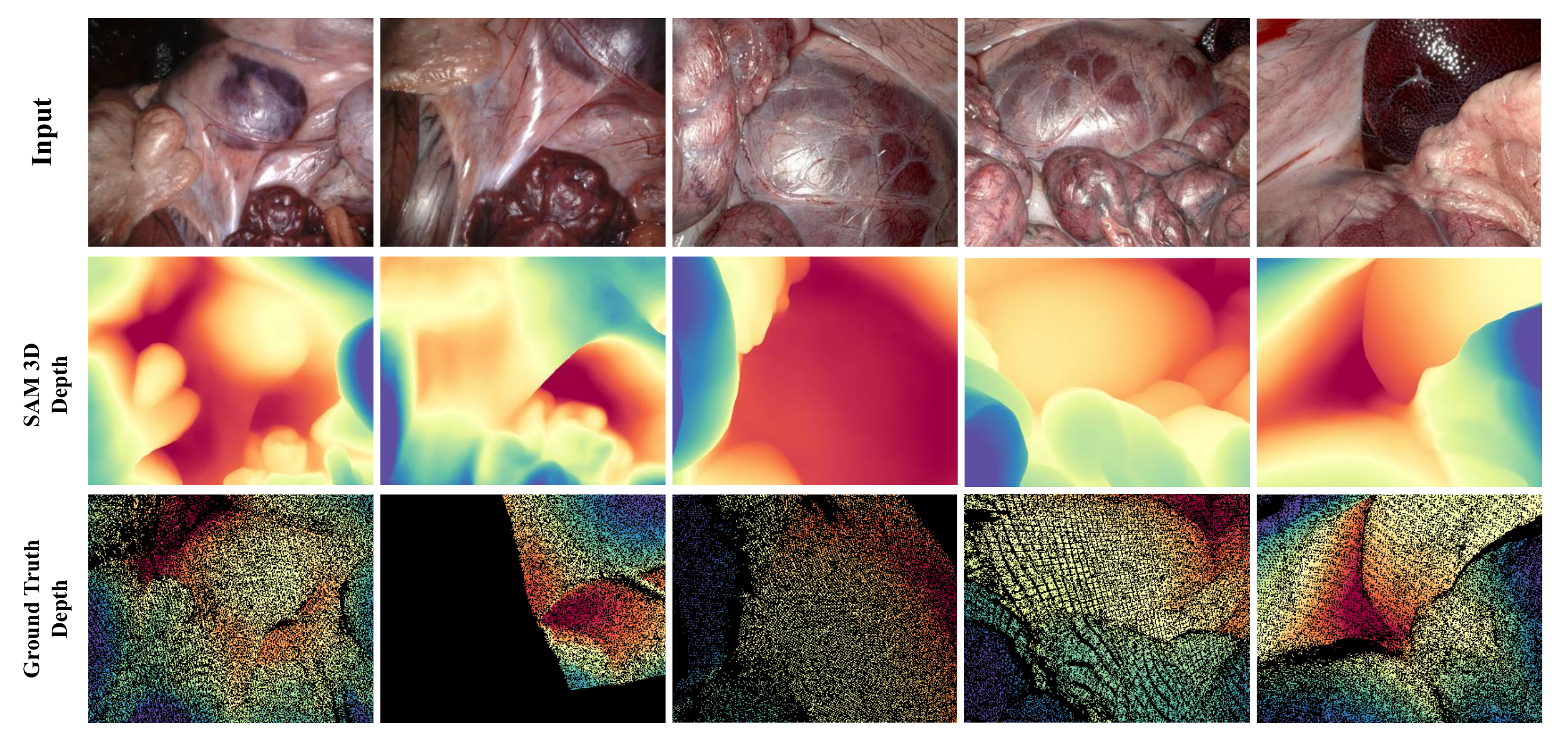}
  \caption{SAM 3D qualitative results of monocular depth estimation.}
  \label{fig:depth}
\end{figure*}

\subsection{Results and Analysis}
On the SCARED dataset, SAM 3D in a zero-shot setting achieves lower error and higher accuracy than several prior methods that were fine-tuned on endoscopic 3D data. The results are impressive, as those baseline models were trained or adapted with large amounts of domain data. The result suggests that SAM 3D’s pretraining provides strong depth cues and good generalization to endoscopic scenes without further tuning. Possible reasons include broad and diverse pretraining or strong implicit priors for shape and depth. This finding highlights SAM 3D’s robustness and calls for careful follow-up to verify and understand the effect in clinical settings.

\section{3D Instrument Segmentation}

\subsection{Implementation Details}

\noindent \textbf{Datasets.} We compare the 3D instrument segmentation results on the public StereoMIS~\cite{hayoz2023learning} and EndoNeRF~\cite{wang2022neural} datasets. StereoMIS is a benchmark dataset for camera pose estimation in dynamic surgical environments captured by the da Vinci Xi surgical system, which includes 16 full-length stereo sequences from 3 porcine and 3 human subjects. EndoNeRF~\cite{wang2022neural} is a dataset for deformable tissue reconstruction, which includes two case-specific 50-100 prostatectomy frames extracted from a single-viewpoint stereo endoscope. Both datasets provide manually labelled binary instrument masks, and we generate the ground truth depths with the stereo images following~\cite{wang2022neural}. For StereoMIS, we select four sequences with the odd-numbered images from 14000-14099 and 15300-15399 of sequence P1, 3700-3800 and 6100-6200 of sequence P2. For EndoNeRF, we test on the whole Pulling and Cutting sequences with 63 and 156 images, respectively.
\\
\\
\noindent \textbf{Evaluation Details.}
We report the IoU, Accuracy (Acc), Precision, Recall, and Dice of the 3D segmentation for evaluation. To calculate the 3D evaluation metrics for depth+semantic prediction, we first reproject both predicted and ground-truth depth maps into camera-space 3D point clouds using the provided intrinsic matrix. Then, we align the predicted depth with the ground truth and attach the semantic labels to obtain point clouds with labels. Next, we perform 1-nearest-neighbor matching in 3D space with a small distance threshold of 10 mm. The metrics are computed over the matched points for the 3D semantic evaluation. All experiments are run on the NVIDIA RTX A6000 GPU.

\begin{table}[t]
\scriptsize
\centering
\caption{Quantitative comparison for 3D surgical instrument segmentation.
}
\vspace{-1em}
\label{tab:3d_seg}
\resizebox{0.8\linewidth}{!}{%
\begin{tabular}{l|ccccc}
\toprule
 & \textbf{IoU}$\uparrow$ & \textbf{Acc}$\uparrow$ & \textbf{Precision}$\uparrow$ &\textbf{Recall}$\uparrow$ & \textbf{Dice}$\uparrow$
\\ 
\midrule

StereoMIS\_P1 & 19.09 & 24.85 & 71.62 & 20.78 & 31.36 \\
StereoMIS\_P2 & 3.16 & 22.20 & 13.63 & 6.81 & 6.06\\
EndoNerf Pulling & 42.60 & 74.50 & 99.15 & 42.79 & 59.27\\
EndoNerf Cutting & 47.79 & 82.80 & 89.55 & 50.70 & 64.35\\

\bottomrule
\end{tabular}
}
\end{table}

\begin{figure*}[t]
  \centering
  \includegraphics[width=0.9\linewidth, trim=0 60 0 0]{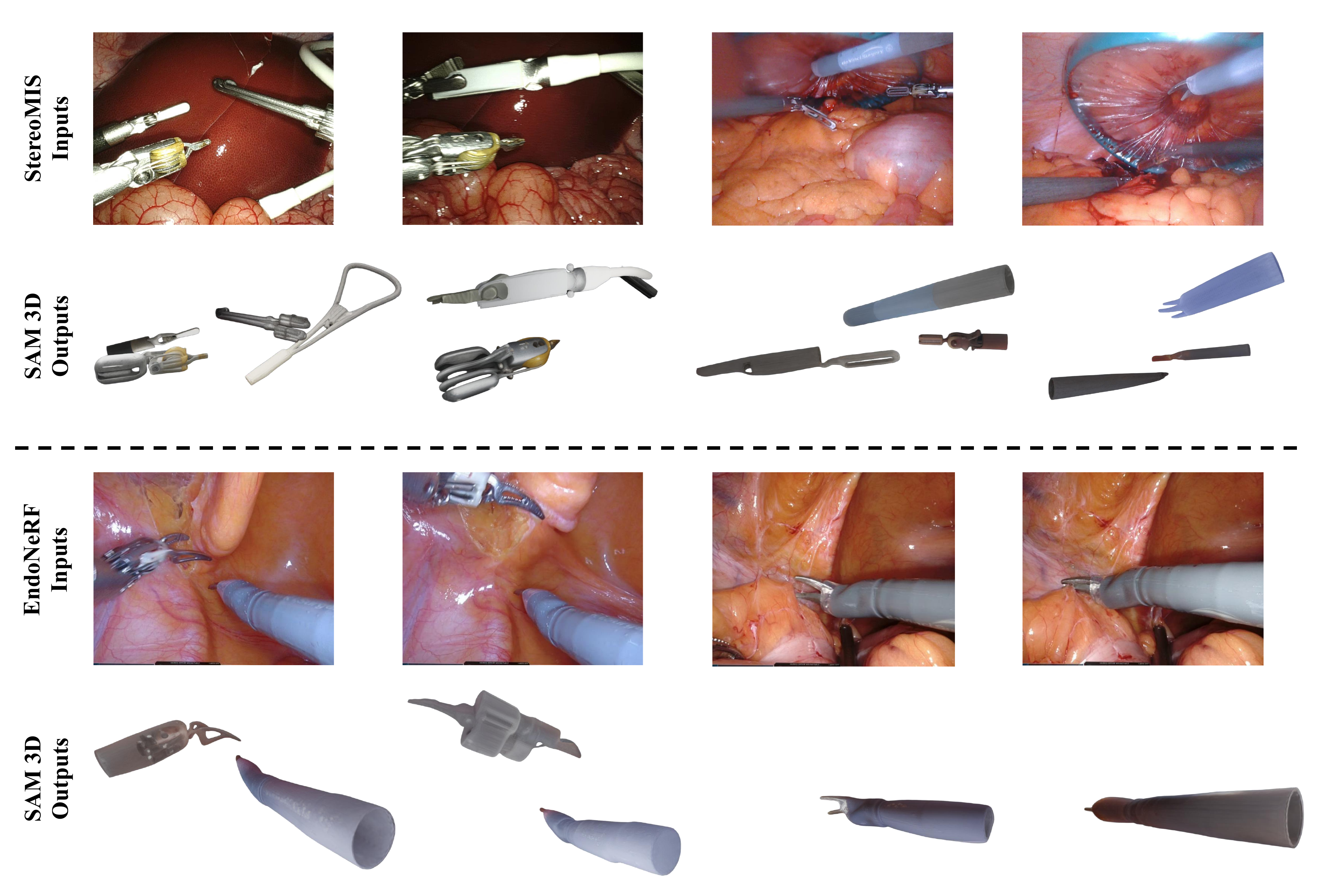}
  \caption{Visualization of the 3D instrument segmentation and reconstruction results using SAM 3D.}
  \label{fig:3dseg}
\end{figure*}

\subsection{Results and Analysis}

Table~\ref{tab:3d_seg} shows the quantitative result for 3D instruments segmentation obtained with SAM 3D on the StereoMIS and EndoNeRF datasets. Across the four surgical sequences, SAM 3D achieves a mean IoU of 28.16 \%, mean accuracy of 51.09 \%, and mean F1-score of 40.2\%. Although the experiment results appear modest, they are obtained in a zero-shot setting—no stereo or temporal pairs were used for training, and no task-specific fine-tuning was performed. The gap is most pronounced on StereoMIS P1 and P2, where rapid instrument motion and narrow-baseline stereo yield severe depth uncertainty, where SAM 3D IoU remains < 20 \%. On the EndoNeRF sequences, where tissue deformation is large but the camera is static, and the depth range is smaller, SAM 3 improves to 42–48 \% IoU, demonstrating that its implicit monocular priors are better suited to controlled camera settings.

We also present the qualitative results, as shown in Fig.~\ref{fig:3dseg}. On the EndoNeRF Pulling sequence, the reconstructed point cloud preserves the shaft curvature and jaw opening of the large needle driver, demonstrating high fidelity close to ground truth. On the Cutting sequence, the bipolar forceps are correctly separated from the deforming tissue surface, and the instrument axis is continuous. In contrast, on StereoMIS P2, the same forceps is fragmented into two disconnected clusters because the rapid out-of-plane motion fails the implicit in SAM 3D’s monocular depth head. Additionally, specular reflections on the metallic shaft further corrupt the predicted depth, leading to corrupted reconstruction.

\section{Conclusion}
\label{sec:conclusion}
In this work, we conducted a comprehensive empirical evaluation of SAM 3 and SAM 3D in robotic surgery across 2D segmentation, video tracking, language-based prompting, and 3D perception. Our experiments show that, under spatial prompts (point and box), SAM 3 achieves strong zero-shot performance on EndoVis17 and EndoVis18, often matching or surpassing prior state-of-the-art methods, making it an effective off-the-shelf tool for label-efficient surgical analysis. In contrast, the newly introduced language-based prompts perform poorly in surgical scenes, revealing a clear semantic and domain gap, yet also pointing to a promising direction for surgery-specific adaptation and finetuning. On the 3D side, SAM 3D demonstrates surprisingly strong pretrained depth priors and plausible 3D instrument reconstruction on endoscopic benchmarks, but still struggles in highly dynamic, narrow-baseline settings and remains computationally expensive for real-time use. Overall, our findings suggest that while SAM 3 is already practical for spatially prompted surgical segmentation, its language interface and 3D perception should be viewed as emerging capabilities that require dedicated surgical data and modeling to fully translate into reliable operating-room tools.

\bibliographystyle{splncs04}
\bibliography{references}

\end{document}